\documentclass[conference]{IEEEtran}
\IEEEoverridecommandlockouts
\usepackage{times}
\usepackage[numbers]{natbib}
\usepackage{multicol}
\usepackage[bookmarks=true]{hyperref}
\usepackage[utf8]{inputenc}
\usepackage{xcolor}
\usepackage{graphicx} 
\usepackage{textgreek}
\usepackage{amsfonts}
\usepackage{amsmath}
\usepackage{amssymb}
\usepackage{subfig}
\usepackage{bbm}
\usepackage[ruled,vlined]{algorithm2e}
\usepackage{tikz}
\usepackage{booktabs}
\usetikzlibrary{shapes.geometric, arrows.meta, positioning, fit, backgrounds, calc}

\definecolor{my_pink}{HTML}{be0027}

\begin{document}

\title{SOP: A Scalable Online Post-Training System for Vision-Language-Action Models}
%%J.L.12.8 do we need the word "high-performance" here? how about "SOP: A scalable online post-training framework for robotic generalist policies"? SOP doesn't necessarily need to restrict to VLAs either
% Mingjie Pan*, Siyuan Feng*, Xinchen Li, Qinglin Zhang, Jianheng Song, Chendi Qu, Yi Wang, Chuankang Li, Ziyu Xiong, Yi Liu, Zhi Chen, Jianlan Luo 

\author{
    Mingjie Pan*$^{1}$ \quad Siyuan Feng*$^{1}$ \quad Qinglin Zhang$^{1}$ \quad Xinchen Li$^{1}$  \quad Jianheng Song$^{1}$ \\Chendi Qu$^{1}$ \quad Yi Wang$^{1,2}$ \quad Chuankang Li$^{1}$ \quad Ziyu Xiong$^{1}$ \quad Zhi Chen$^{1}$ \quad Yi Liu$^{1}$ \quad Jianlan Luo$^{1,2\dagger}$
    \thanks{$*$ Equal contribution.} 
    \thanks{$\dagger$ Corresponding author.}  \\ 
    $^1$Agibot Research. $^2$Shanghai Innovation Institute. \\
    
    \\
\href{https://www.agibot.com/research/sop}{\textcolor{my_pink}{https://www.agibot.com/research/sop}
}
}

% % }
% \author{
% Charles Xu$^{1,\dagger}$ \quad Qiyang Li$^{1}$ \quad Jianlan Luo$^{1,\dagger}$ \quad Sergey Levine$^{1}$ \\
% $^1$University of California, Berkeley. $\dagger$ Project Leads \\ Emails: \texttt{\{xuc, qcli\}@berkeley.edu, \{jianlanluo, svlevine\}@eecs.berkeley.edu}}

\maketitle

\begin{abstract}
Vision-language-action (VLA) models achieve strong generalization through large-scale pre-training, but real-world deployment requires expert-level task proficiency in addition to broad generality. Existing post-training approaches for VLA models are typically offline, single-robot, or task-specific, limiting effective on-policy adaptation and scalable learning from real-world interaction.
We introduce a Scalable Online Post-training (SOP) system that enables online, distributed, multi-task post-training of generalist VLA models directly in the physical world. SOP tightly couples execution and learning through a closed-loop architecture in which a fleet of robots continuously streams on-policy experience and human intervention signals to a centralized cloud learner, and asynchronously receives updated policies. This design supports prompt on-policy correction, scales experience collection through parallel deployment, and preserves generality during adaptation. SOP is agnostic to the choice of post-training algorithm; we instantiate it with both interactive imitation learning (HG-DAgger) and reinforcement learning (RECAP). Across a range of real-world manipulation tasks including cloth folding, box assembly, and grocery restocking, we show that SOP substantially improves the performance of large pretrained VLA models while maintaining a single shared policy across tasks. Effective post-training can be achieved within hours of real-world interaction, and performance scales near-linearly with the number of robots in the fleet. These results suggest that tightly coupling online learning with fleet-scale deployment is instrumental to enabling efficient, reliable, and scalable post-training of generalist robot policies in the physical world.
\end{abstract}

\IEEEpeerreviewmaketitle

\section{Introduction}

\emph{In the long history of humankind, those who learned to collaborate and improvise most effectively have prevailed.}

\vspace{0.3em}
\hrule
\vspace{0.3em}
\hfill Charles Darwin, \emph{The Descent of Man}

\vspace{1em}

\begin{figure}[t]
    \centering
    \includegraphics[width=0.8\linewidth]{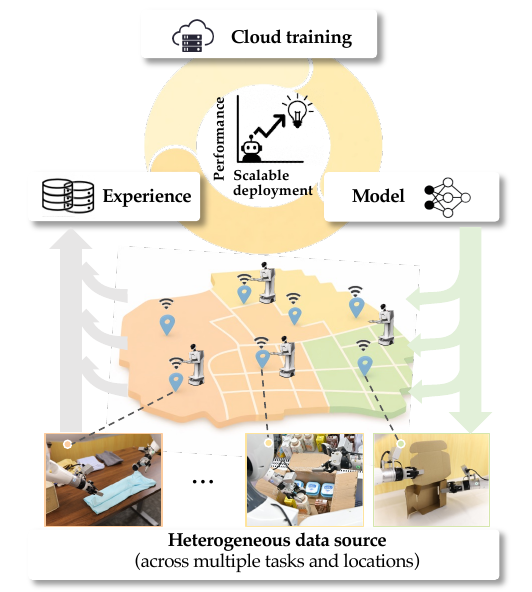}
    \caption{\textbf{Scalable Online Post-training (SOP).} A fleet of robots continuously collects experience across diverse tasks, streams interaction data to a centralized cloud server, and receives updated control policies asynchronously—enabling VLA models to improve proficiency on each task while preserving generality.}
    \label{fig:teaser}
\end{figure}

Deploying general-purpose robots at scale is becoming increasingly tractable~\cite{zitkovich2023rt,black2024pi_0,kim2024openvla,intelligence2025pi_,liu2025hybridvla}. However, generality alone does not suffice for real-world deployment. Real-world deployment demands \emph{high-performance generalists}—systems that not only generalize across diverse tasks, but also achieve expert-level proficiency when instantiated in any specific setting.
Consider a household robot: it must fold laundry, organize shelves, and assemble furniture, while exhibiting the reliability and precision expected of a dedicated appliance. 
Neither generality nor proficiency in isolation meets this bar; the two capabilities must coexist within a single system.

Vision-language-action (VLA) models represent substantial progress toward providing the generality component of this requirement~\cite{zitkovich2023rt, octo_2023, kim2024openvla, black2024pi_0,intelligence2025pi_}. By unifying visual perception, language understanding, and action generation within a single architecture, VLAs trained on Internet-scale data exhibit very strong generalization capability across tasks, objects, and embodiments~\cite{brohan2022rt, zitkovich2023rt, gr00tn1openfoundation}. 
The remaining challenge is how to endow these models with expert-level proficiency without sacrificing this carefully achieved generality.

The answer lies in \emph{post-training}—adapting the pre-trained model to specific downstream deployment scenarios~\cite{ouyang2022rlhf, amin2025pi,gr-rl,kai0}.
In domains such as large language models (LLMs), post-training via online reinforcement learning (RL) and human feedback has proven to be very effective~\cite{ouyang2022rlhf, deepseek-r1}, enabling models to continuously improve through large-scale distributed training~\cite{yao2023deepspeed, hu2024openrlhf}. Yet system-level realizations of online learning with distributed data collection remain largely unexplored for VLA post-training in the physical world.

Despite substantial progress in algorithmic post-training methods, existing VLA approaches remain fundamentally constrained by the absence of a unified system that couples distributed robot fleets with centralized, online learning. Consequently, most prior work operates in offline, single-robot, and task-specific regimes~\cite{chi2024diffusionpolicy, zhao2023aloha, black2024pi_0, zitkovich2023rt}, where data collection and policy improvement are structurally decoupled. In this setting, offline training on pre-collected demonstrations inevitably suffers from distribution shift, with small execution errors compounding over long horizons. Iterative imitation learning methods, such as DAgger, partially mitigate this issue by incorporating human corrections~\cite{ross2011reduction, laskey2016shiv, laskey2017dart, jang2022bc, hu2025rac}, but their batch-based update cycles introduce latency between execution and correction, limiting their effectiveness in real-time sequential decision making—even in fully online variants such as HG-DAgger~\cite{ kelly2019hg}. 
This observation is consistent with theoretical results highlighting the importance of timely, on-policy corrections for mitigating distribution shift~\cite{kakade2002approximately, agarwal2021theory}.
These limitations are further exacerbated by single-robot data collection, which restricts experience diversity and learning speed, and by task-specific fine-tuning, which often trades generality for  gains in proficiency~\cite{chi2024diffusionpolicy, zhao2023aloha, kim2024openvla,octo_2023}. 
Collectively, these challenges reflect limitations of the underlying learning setting rather than shortcomings of individual algorithms. Consequently, no existing VLA post-training approach simultaneously supports timely on-policy correction, scalable experience collection, and multi-task adaptation within a single generalist model.

To address these challenges, we introduce a Scalable Online Post-training (SOP) system for post-training generalist VLA models directly in the physical world using large-scale real-world interaction.
The key insight is that tightly coupling learning and execution yields a unified feedback loop that enables timely on-policy correction, scales exploration through parallel experience, and preserves generality during adaptation. SOP realizes this insight via a closed-loop architecture in which a robot fleet exchanges on-policy trajectories and human intervention signals with a centralized cloud learner. This collect–train–deploy loop enables low-latency adaptation and scales naturally with fleet size.

While SOP is in principle agnostic to the choice of post-training algorithm, we instantiate it with HG-DAgger~\cite{kelly2019hg} and RECAP~\cite{amin2025pi}, representing interactive imitation learning and reinforcement learning, respectively.
SOP substantially improves the performance of pretrained VLA models across a diverse set of real-world manipulation tasks, including cloth folding, box assembly, and shelf restocking.
Crucially, these gains do not come at the expense of generality: a single model is jointly post-trained across all tasks, with shelf restocking involving a large and diverse set of objects.

Perhaps most surprisingly, SOP enables effective post-training of large VLA models directly in the real world on the order of hours, rather than extended multi-day training cycles. This efficiency arises from prompt on-policy correction that targets the deployed policy’s failure modes, amplified by distributed experience collection. Additionally, SOP exhibits near-linear scaling: increasing the number of robots reduces the wall-clock time required to reach a target performance level approximately proportionally.
Across both HG-DAgger and RECAP, SOP consistently outperforms their non-SOP counterparts, often achieving 2× or greater improvements in success rate, with several tasks approaching near-perfect performance and substantially higher throughput.
In long-horizon evaluations, tasks such as laundry folding and box assembly run continuously for over 36 hours without degradation, demonstrating that SOP provides systematic benefits beyond algorithmic post-training alone.

Our primary contribution is SOP, the first framework for online, distributed, multi-task post-training of VLA models in the physical world. SOP enables a fleet of robots to continuously share real-world experience through a centralized learner, allowing models to rapidly improve task proficiency without sacrificing generality. Crucially, we show that existing post-training algorithms, when instantiated within SOP, can reliably and efficiently improve large pretrained VLA models on challenging dexterous manipulation tasks using only limited real-world interaction.

More broadly, SOP represents a concrete step toward scalable robot learning through shared experience across robot fleets.
By tightly coupling deployment and learning, SOP enables a collection of robots to jointly maintain and refine a continuously evolving VLA directly from real-world interaction.
This coupling establishes a feedback loop in which scaling robot fleets not only improves the efficiency of post-training, but also increases the diversity and relevance of experience available for learning—supporting continual adaptation and robust performance in long-horizon real-world deployments.
These results suggest that fleet-scale deployment can serve as an important and complementary enabler of progress in robot learning systems, alongside advances in algorithms and data.

% Our primary contribution is SOP, the first framework for online, distributed, multi-task post-training of VLA models in the physical world. SOP enables a fleet of robots to continuously stream experience across diverse tasks to a centralized cloud learner, allowing models to improve task proficiency through real-world interaction without sacrificing generality. Crucially, we demonstrate for the first time that, when instantiated within SOP, existing post-training methods can rapidly and reliably improve large pretrained VLA models on a range of challenging dexterous manipulation tasks using only a short amount of real-world interaction—while preserving the broad generalization acquired during pre-training.

% Looking forward, we believe SOP provides a concrete step toward scalable robot learning through shared experience across robot fleets. By enabling robots to jointly maintain and refine a continuously evolving VLA from real-world interaction, SOP supports models that can be broadly deployed and incrementally improved without sacrificing generality. More broadly, SOP implies a positive feedback loop between deployment and learning, where scaling robot fleets enables more effective post-training, which in turn supports broader deployment—complementing advances in model scale and pre-training in other domains.

\section{Related Works}
\begin{figure*}[ht]
    \centering
    \includegraphics[width=\linewidth]{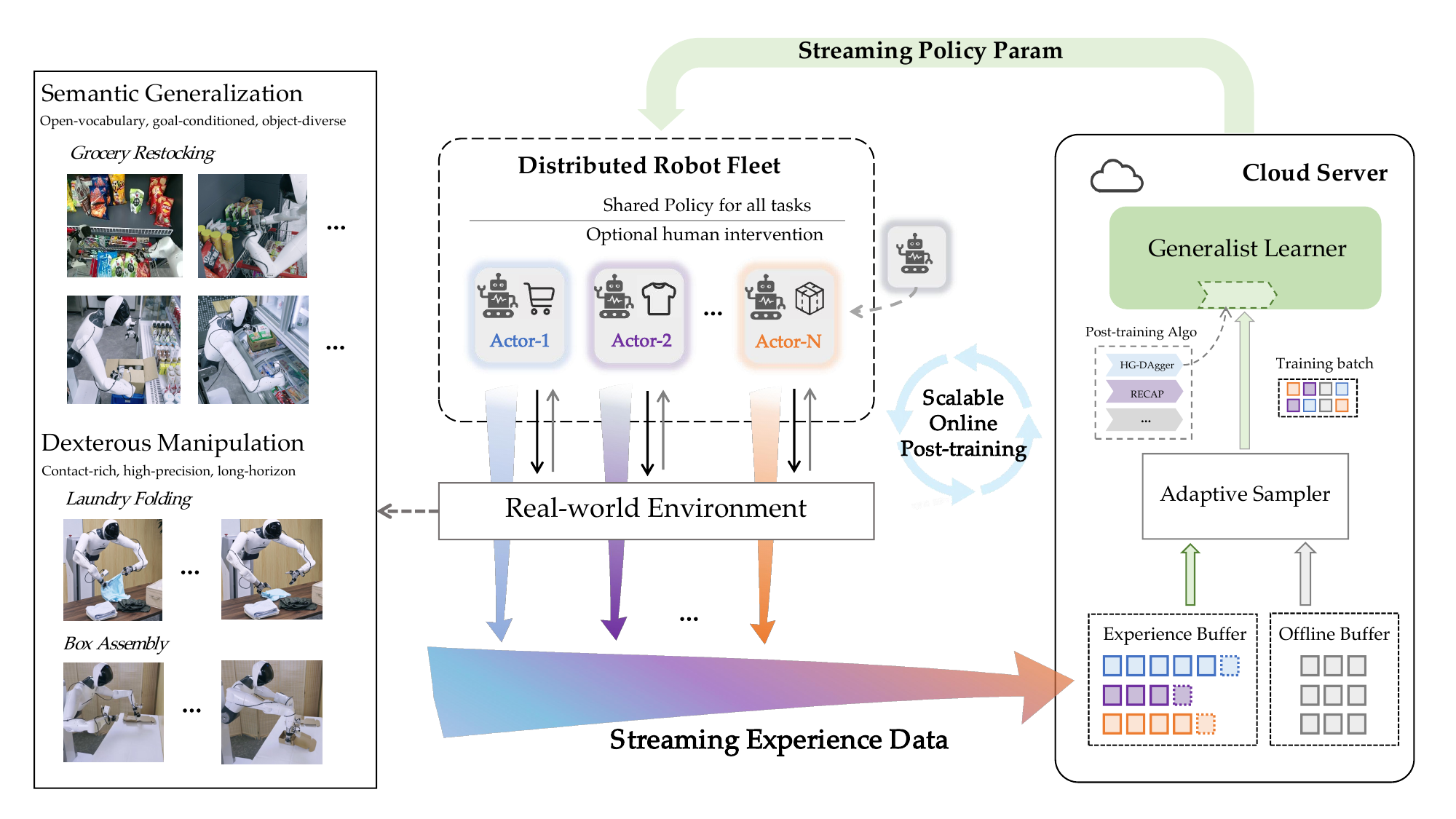}
    \caption{SOP overview. SOP is a scalable actor–learner framework for online, multi-task post-training of generalist policies. The robot fleet streams on-policy rollouts to the cloud learner. Optional human interventions are triggered in failure or uncertain cases, providing corrected trajectories or actions that are incorporated into the streaming experience buffer. The cloud learner constructs task-balanced updates by mixing an online buffer with a static offline buffer, applies a plug-in post-training module (e.g., HG-DAgger/RECAP), and asynchronously broadcasts refreshed weights back to all actors to close a low-latency online training loop.}
    \label{fig:method}
\end{figure*}
%%J.L.12.8 we need motivation here, something along the line "our method combines xxx and yyy to do zzz, thus we survey xxx and zzz"
SOP is a fundamentally integrative system framework: by combining online 
learning, distributed data collection, and multi-task training, it addresses 
the limitations of existing VLA post-training methods. 
We thus survey related works along these three axes.

\subsection{VLA Post-training}
VLA models achieve broad generalization through large-scale pre-training on diverse multimodal data~\cite{brohan2022rt,zitkovich2023rt,kim2024openvla,octo_2023,intelligence2025pi_,bjorck2025gr00t,tri2025lbm, agibotworld, liu2025unified, shukor2025smolvla}.
To adapt these models to deployment settings, post-training methods broadly fall into two categories: supervised finetuning and reinforcement learning.
Supervised finetuning adapts VLA models using task-specific demonstrations~\cite{black2024pi_0,liu2024rdt,kim2024openvla}. While stable and effective, these methods rely on static datasets, limiting their ability to address distribution shift or to improve beyond demonstration coverage.
Reinforcement learning (RL)~\cite{sutton1998reinforcement} improves policies through environment interaction and feedback, encompassing both online and offline RL~\cite{lange2012batch, levine2020offline}.
Online RL algorithms such as PPO~\cite{schulman2017proximal} and GRPO~\cite{grpo2024} have shown strong results in domains including robotic locomotion and LLMs, but face practical challenges in real-world robotic manipulation, including high variance and training instability~\cite{lu2025vla,liu2025can,li2025simplevla,chen2025pi_,2025zhaivlac}.
Behavior-regularized RL methods can improve stability~\cite{chen2025conrft, lei2025rl}, but often bias asymptotic performance and are typically task-specific, limiting applicability to generalist settings.
Among offline RL approaches~\cite{nair2020awac, kostrikov2021offline, wang2020critic}, RECAP~\cite{amin2025pi} is most closely related. RECAP combines reward feedback with human interventions through iterative offline training, focusing on task-specific specialization. While effective for individual tasks, it does not directly support continuous online improvement of a single generalist policy across tasks.
RLDG \cite{xu2024rldg} adopts a complementary strategy by first using task-specific RL to generate high-quality trajectories, which are subsequently distilled into a generalist policy via behavior cloning. While this distillation step helps retain generality, data generation in RLDG remains offline, single-robot, and requires training separate RL policies for each task, limiting scalability.
In contrast, SOP directly updates a generalist policy through continuous online learning with distributed, multi-task data collection, enabling performance improvement while preserving generality.
% \mj{without citing recent offline/single-actor post-training paper like GR-RL\cite{gr-rl}(bytedance) and kai0\cite{kai0}(mmlab)}

\subsection{Interactive and Online Learning}
Online learning addresses distribution shift by training on states encountered by the current policy, rather than relying solely on pre-collected expert data.
Interactive imitation learning methods such as DAgger~\cite{ross2011reduction} iteratively collect states from the learned policy and query expert labels, while HG-DAgger~\cite{kelly2019hg} shifts to real-time intervention where the expert only takes over when the policy is about to fail, reducing human burden while directly correcting on-policy mistakes.
Sample-efficient online RL methods leverage demonstrations to accelerate learning: RLPD~\cite{ball2023efficient} mixes offline data with online experiences, SERL~\cite{luo2024serl} combines demonstrations with online RL for robotic tasks, and HIL-SERL~\cite{luo2025precise} further integrates human intervention, achieving expert-level performance on complex long-horizon tasks. However, these methods remain \emph{single-robot} (limiting data collection throughput) and \emph{task-specific} (requiring separate training for each task). SOP extends online learning to distributed multi-task settings, enabling parallel state-space coverage across robot fleets while preserving generality.

\subsection{Distributed and Multi-task Robot Learning}
Scaling beyond single-robot data collection requires distributed architectures, while multi-task learning prevents overfitting to narrow objectives.
Distributed RL systems such as Gorila~\cite{d4pg}, A3C~\cite{mnih2016asynchronous}, and IMPALA~\cite{espeholt2018impala} pioneered actor-learner architectures for accelerated training, but target simulation environments where reset is trivial and do not address real-world deployment with human oversight. Fleet-DAgger~\cite{hoque2023fleet} is the closest prior work, establishing a multi-robot interactive learning system with scalable human supervision. It achieves online distributed learning, but remains simulation-only, not designed for large VLA models, and limited to single-task learning. Multi-task robot learning~\cite{singh2020scalable,kalashnikov2021mt,mandi2022cacti} demonstrates the benefits of sharing experience across tasks, but these methods do not combine online learning with distributed collection for VLA post-training.
In summary, SOP is the first framework that simultaneously enables online, distributed, multi-task post-training for generalist models in the physical world.

\section{Preliminaries and Problem Statement}

\subsection{Preliminaries} 
% In this paper, we leverage an online fine-tuning which relies on the robot interacting within its own environment. We formulate this interaction with a Markov decision process (MDP). A MDP is defined as a tuple $\mathcal{M}=(\mathcal{S},\mathcal{A},T,r,\gamma)$, where $\mathcal{S}$ is the state space, $\mathcal{A}$ is the action space (discrete or continuous, potentially chunked), $T(s'|s,a)$ and $r(s,a)$ denote the transition kernel and environment reward at $(s,a)$ and $\gamma \in (0,1]$ is the discount factor.
% %, and $\Omega,\mathcal{O}(o|s)$ are observation space and observation function mapping a state $s$ to $o$. 
% For VLA training, the state $s$ is usually composed of vision images, language prompt and robot joint poses. 
% Consider a policy $\pi_{\theta}(a|s)$ parameterized by $\theta$, mapping from $\mathcal{S}$ to $\mathcal{A}$. At time $t$, the agent executes $a_t \sim \pi_{\theta}(\cdot|s_t)$ and transfers to a new state $s_{t+1} \sim T(\cdot|s_t,a_t)$. The environment then gives a reward feedback $r_t$. %  We record this transition as $(o_t, a_t, r_t, o_{t+1})$.
We formulate the considered robot control problem as a Markov decision process (MDP) $\mathcal{M}=(\mathcal{S},\mathcal{A},T, r, \gamma)$, where $\mathcal{S}$ is the state space, $\mathcal{A}$ is the action space, $T(s'|s,a)$ and $r(s,a)$ denote the transition dynamics and environment reward at $(s,a)$ and $\gamma \in (0,1]$ is the discount factor. For VLA models, the state $s$ typically comprises visual observations, language instructions, and robot proprioceptive information. A policy $\pi_{\theta}(a|s)$ defines a distribution over actions given state $s$; at each step, the agent samples $a \sim \pi_{\theta}(\cdot|s)$ and transitions to $s' \sim T(\cdot|s,a)$.
%Importantly, SOP \qcd{don't think SOP should be pointed out here} treats the policy and its action parameterization as a plug-and-play component; the system is agnostic to the specific modeling choice (e.g., discrete vs.\ continuous, diffusion/flow-based, etc.).

\subsection{Problem Statement}
Consider a distributed robot system. $N$ robots are deployed in various environments, operating different tasks. We use a domain variable $\phi$ to model the heterogeneity, which induces a family of MDPs $\mathcal{M}(\phi)$. Specifically, the $i$-th robot's interaction is governed by $\mathcal{M}^i := \mathcal{M}(\phi_i)$, where $\phi_i \sim p(\phi)$ for $i= 1, 2,\dots,N$.  
The goal of post-training is to adapt a pretrained base policy $\pi_{\theta_0}$ to each deployment domain by leveraging online interaction data collected from the robot fleet. 
%In our experiments, $\pi_{\theta_0}$ corresponds to a VLA base policy pretrained for one round on our in-house robot dataset (Appendix~\ref{app:impl}).
This usually involves a multi-round optimization procedure. At the $k$-th iteration, robots execute the current policy $\pi_{\theta_k}$ and collect trajectories, including both autonomous rollouts and potential human interventions, into a dataset $\mathcal{D}_k$. The policy is then updated by minimizing a post-training objective defined over collected samples:
\begin{equation}
\theta_{k+1} = \arg \min_{\theta} \mathbb{E}_{(s,a) \sim \mathcal{D}_k} \mathcal{L}_{PT}(\pi_{\theta}; s,a).
\end{equation}
$\mathcal{L}_{PT}$ denotes the post-training loss corresponding to the specific chosen post-training algorithm $\mathcal{G}$, which can be formulated as log-likelihood loss or diffusion/flow-based loss.

\section{Scalable Online Post-training}

We present \emph{Scalable Online Post-training} (SOP), a closed-loop actor--learner framework for adapting a pretrained VLA policy using continual real-world interaction from a heterogeneous robot fleet. SOP consists of (i) distributed on-policy data collection by robot actors, (ii) centralized cloud optimization on mixed online and offline data, and (iii) low-latency model synchronization back to actors. Importantly, SOP is \emph{algorithm-agnostic}: it specifies the system-level dataflow and synchronization, while the concrete parameter update method can be replaced by any post-training algorithm. In this paper, we instantiate SOP with two existing post-training methods---HG-DAgger~\cite{kelly2019hg} and RECAP~\cite{amin2025pi}---and show that SOP \emph{upgrades} them into practical on-policy, online post-training by continuously streaming fresh experience and applying frequent asynchronous model updates.

\begin{algorithm}[t]
\caption{Scalable Online Post-training for VLA}
\label{alg:distributed_vla}
\KwIn{
Initial policy $\pi_{\theta_0}$,
Offline buffer $\mathcal{B}_{\text{off}}$,
Post-training algorithm $\mathcal{G}$,
Adaptive sampler $\mathcal{S}$
}
Initialize online buffer $\mathcal{B}_{\text{on}} \leftarrow \emptyset$\;
Broadcast $\pi_{\theta_0}$ to all actors;

\textbf{Actor $i$ (parallel):}\\
\While{acting}{
    Execute policy $\pi_\theta$\ and collect rollouts $\tau_{\pi}^i$;\\
    \If{human has control}{Collect interventions $\tau_{H}^i$;}
    Upload $\tau_{\pi}^i \cup \tau_{H}^i$ to $\mathcal{B}_{\text{on}}$.
}
%\BlankLine
\textbf{Cloud Learner (asynchronous):}\\
\While{training}{
    Sample a training batch $\xi \leftarrow \mathcal{S}(\mathcal{B}_{\text{on}} \cup \mathcal{B}_{\text{off}})$;\\
    %Sample $\mathcal{D}_{\text{on}} \leftarrow \mathcal{S}(\mathcal{B}_{\text{on}})$ and $\mathcal{D}_{\text{off}} \leftarrow \mathcal{S}(\mathcal{B}_{\text{off}})$\;
    Update policy parameters:
    $\theta \leftarrow \mathcal{G}(\theta, \xi)$\;
    
    Stream updated policy $\pi_\theta$ to all actors;
}

%\BlankLine
\end{algorithm}

\subsection{Algorithm Framework}
Algorithm~\ref{alg:distributed_vla} summarizes SOP. We start from a pretrained policy $\pi_{\theta_0}$ (see more details in Appendix~\ref{app:impl}) and broadcast it to all $N$ robot actors. Each actor $i$ continuously executes the latest available policy $\pi_{\theta}$ in its local domain $\mathcal{M}^i$ and uploads trajectories to a shared online experience buffer in parallel. Trajectories include autonomous rollouts $\tau_{\pi}^i$ and, when available, human interventions $\tau_H^i$ for correction. At the same time, a centralized cloud learner continuously samples training batches from a mixture of the online buffer and static offline buffer and updates the shared parameters via a post-training algorithm $\mathcal{G}$. Updated parameters are then streamed back to all actors asynchronously.

%Suppose we have a base policy $\pi_{\theta_0}$ trained on the prior data. $\pi_{\theta_0}$ is shared with all $N$ robots at the beginning of the post-training process. At time $t$, we define the robot $i$ executes the policy $\pi_{\theta^i(t)}$ and collects rollout episodes.
%We define a model deployment cycle. At $k$-th deployment cycle, the robot $i$ executes the policy $\pi^i_{\theta_k}$ and collect a rollout episode $\tau^i_k=(s^i_{k,0}, a^i_{k,0}, s^i_{k,1}, a^i_{k,1}, s^i_{k,2},\dots)$.

%Each episode is augmented with comprehensive metadata, including the natural language prompt, reward and completion signals, human intervention markers, and a unique identifier. Upon episode termination, an asynchronous streamer compresses and uploads the episode package to an Object Storage Service. The uploaded data are then appended into a cloud-hosted ring buffer, which serves as the ingestion interface to the centralized learner.
Specifically, let $\tau^i(t)$ denote all episodes uploaded to the cloud by robot $i$ up to wall-clock time $t$. The aggregated online experience available by time $t$ is
$\mathcal{B}_{\text{on}}(t) = \cup_{i=1}^N \tau^i(t)$.
%On the cloud, the ring buffer is exposed through a dataset abstraction, coupled with a dedicated adaptive sampler.  We have the following training dataset at each step.
At training step $j$ (wall-clock time $t_j$), the training batch sampled by the cloud learner is denoted as
\begin{equation}
\xi_j := \mathcal{S}_j(\mathcal{B}_{\text{on}}(t_j) \cup \mathcal{B}_{\text{off}}),
\end{equation}
where $\mathcal{S}_j$ is a designed adaptive data sampler that enables dynamic online/offline composition via configurable weighting schemes (see the design details in Sec. \ref{adaptive_sampler}). $\mathcal{B}_{\text{off}}$ is a static offline buffer containing prior human demonstrations.
With $\xi_j$, the model parameter is updated by the learner through
\[
\theta \xleftarrow{} \arg 
\min_{\theta} \mathbb{E}_{(s,a) \sim \xi_j} \mathcal{L}_{PT}(\pi_{\theta}; s,a).
\]
Then the latest model parameter $\theta$ is streamed to all actors.

\subsection{System Infrastructure}
We develop a distributed actor--learner data infrastructure designed for real-world robot fleets, as illustrated in Figure~\ref{fig:infra-arch}. Each robot actor runs an edge-side client that buffers episodes locally and uploads them asynchronously to object storage at episode boundaries. Uploaded episodes are then appended into a cloud-hosted online buffer, which the learner consumes independently via notifications and on-demand retrieval.

To close the loop, updated model parameters are synchronized from the cloud learner to robot actors through a lightweight publish--subscribe channel at short intervals. Actors fetch the latest checkpoints with end-to-end latencies typically on the order of seconds to tens of seconds (scaling with model size) and apply updates at safe boundaries (e.g., between episodes), preventing mid-episode policy changes from corrupting logged trajectories. This decoupling allows actors and learners to scale independently and remains robust to transient network disruptions.
See more design details in Appendix \ref{app-infra}.

\subsection{Adaptive Sampling Strategy}
\label{adaptive_sampler}
To preserve multi-task coverage while adapting quickly to newly collected on-policy data, we use a task-balanced adaptive sampling strategy $\mathcal{S}_j$ at learner step $j$. Assume the training data are partitioned into $M$ tasks indexed by $m \in \{1,2,\dots,M\}$. At the inter-task level, we enforce uniform task weights $\omega^m = 1/M$ so that each task contributes equally. At At the intra-task level, for task $m$, we adjust the sampling ratio between the task's online buffer $\mathcal{B}_{\text{on}}^m$ and offline buffer $\mathcal{B}_{\text{off}}^m$ based on recent training losses.
%The training data are partitioned into $M$ tasks. At the inter-task level, we enforce uniform task weights $\omega_m = 1/M$ so that each task contributes equally. At the intra-task level, we dynamically adjust the sampling ratio between online buffer $\mathcal{D}^{\text{on}}_m$ and offline buffer $\mathcal{D}^{\text{off}}_m$ based on recent training losses.

% Specifically, for each task we maintain sliding-window estimates of the online and offline losses with a window size $W$ (default $W=200$) as $\bar{l}^{\text{on}}_m = \frac{1}{W} \sum_{i=j-W}^{j-1} l^{\text{on}}_{m,i}, \bar{l}^{\text{off}}_m = \frac{1}{W} \sum_{i=j-W}^{j-1} l^{\text{off}}_{m,i}$. We compute the online sampling ratio $\omega^{\text{on}}_m$ by a temperature-controlled softmax over the losses:
For each task, we maintain sliding-window estimates of online and offline losses with window size $W=200$: $\bar{l}_{\text{on}}^m = \frac{1}{W} \sum_{i=j-W}^{j-1} l_{\text{on}}^{m,i}$ and $\bar{l}_{\text{off}}^m = \frac{1}{W} \sum_{i=j-W}^{j-1} l_{\text{off}}^{m,i}$. The online sampling ratio $\omega_{\text{on}}^m$ is computed by:

\begin{equation}
\omega_{\text{on}}^m = \frac{\exp(\alpha \cdot \bar{l}_{\text{on}}^m)}{\exp(\alpha \cdot \bar{l}_{\text{on}}^m) + \exp(\bar{l}_{\text{off}}^m)}
\end{equation}
\iffalse
\begin{equation}
\omega^{\text{on}}_m = \frac{\exp(\frac{\alpha \cdot \bar{l}^{\text{on}}_m}{l_{min}}- 1)}{\exp(\frac{\alpha \cdot \bar{l}^{\text{on}}_m}{l_{min}}- 1) + \exp(\frac{ \bar{l}^{\text{off}}_m}{l_{min}}- 1)}
\end{equation}
\fi
% where %and $l_{min} = \min (\alpha \cdot \bar{l}^{\text{on}}_m, \bar{l}^{\text{off}}_m)$.
% $\alpha >1$ is a boost factor that encodes a prior preference for emphasizing online data to accelerate adaptation under distribution shift. To avoid degenerate allocations and preserve both adaptation and coverage, we further clip the online ratio to a bounded interval:
% \[
% \omega^{\text{on}}_m \xleftarrow{} \text{clip} (\omega^{\text{on}}_m, \underline{\omega} = 0.2, \bar{\omega} = 0.8).
% \]
% Then, conditioned on a task $m$, we sample training data from $\mathcal{D}^{\text{on}}_m$ with probability $\omega^{\text{on}}_m$ and from $\mathcal{D}^{\text{off}}_m$ with probability $1-\omega^{\text{on}}_m$. This design guarantees equal task coverage while enabling a loss-driven, per-task adaptation of the online/offline mixture.
where $\alpha>1$ is a boost factor that prioritizes online data to accelerate adaptation under distribution shift. To avoid extreme allocations, we clip $\omega_{\text{on}}^m$ to the interval $\begin{bmatrix}
    0.2, 0.8
\end{bmatrix}$.
Given task $m$, we sample from $\mathcal{B}_{\text{on}}^m$ with probability $\omega_{\text{on}}^m$ and from $\mathcal{B}_{\text{off}}^m$ with probability $1-\omega_{\text{on}}^m$. This sampling strategy ensures equal task coverage while enabling loss-driven adaptation of the online/offline data mixture.

\subsection{Post-training Learning Module}

SOP decouples the \emph{system} (distributed dataflow and synchronization) from the \emph{algorithm} (how to update $\theta$ from a batch). This is captured by the post-training module $\mathcal{G}$ in Algorithm~\ref{alg:distributed_vla}. Any existing post-training method that consumes logged experience and returns updated parameters can be plugged into SOP. Below we summarize the original characteristics of HG-DAgger and RECAP, and describe how SOP turns each into an on-policy, online post-training procedure via continuous data streaming and asynchronous updates.

\paragraph{\textbf{HG-DAgger}}
HG-DAgger~\cite{kelly2019hg} is an interactive imitation learning method in which a human supervisor provides real-time interventions when the robot is about to fail, yielding corrective supervision on hard, on-policy states with reduced human effort compared to full teleoperation. In SOP, these intervention segments (together with autonomous rollouts and offline demonstrations) are continuously streamed into the shared buffer and consumed by the cloud learner for frequent asynchronous updates. This turns HG-DAgger into a practical, fleet-scale \emph{on-policy online} post-training procedure by reducing the latency between failure, correction, and model update.

\paragraph{\textbf{RECAP}}
RECAP~\cite{amin2025pi} is an offline RL method for post-training large VLA policies, designed to improve a policy from experience (including autonomous rollouts and optional human corrections). In its standard usage, RECAP is applied in an iterative offline loop (collect experience, train offline, redeploy). SOP makes this workflow \emph{online} by continuously incorporating freshly collected trajectories from the latest deployed policy into the buffer and running RECAP-style updates asynchronously on the evolving dataset. This reduces policy--data staleness and enables continual, on-policy improvement while keeping RECAP itself unchanged.

\begin{figure*}
\centering
\includegraphics[width=.9\linewidth]{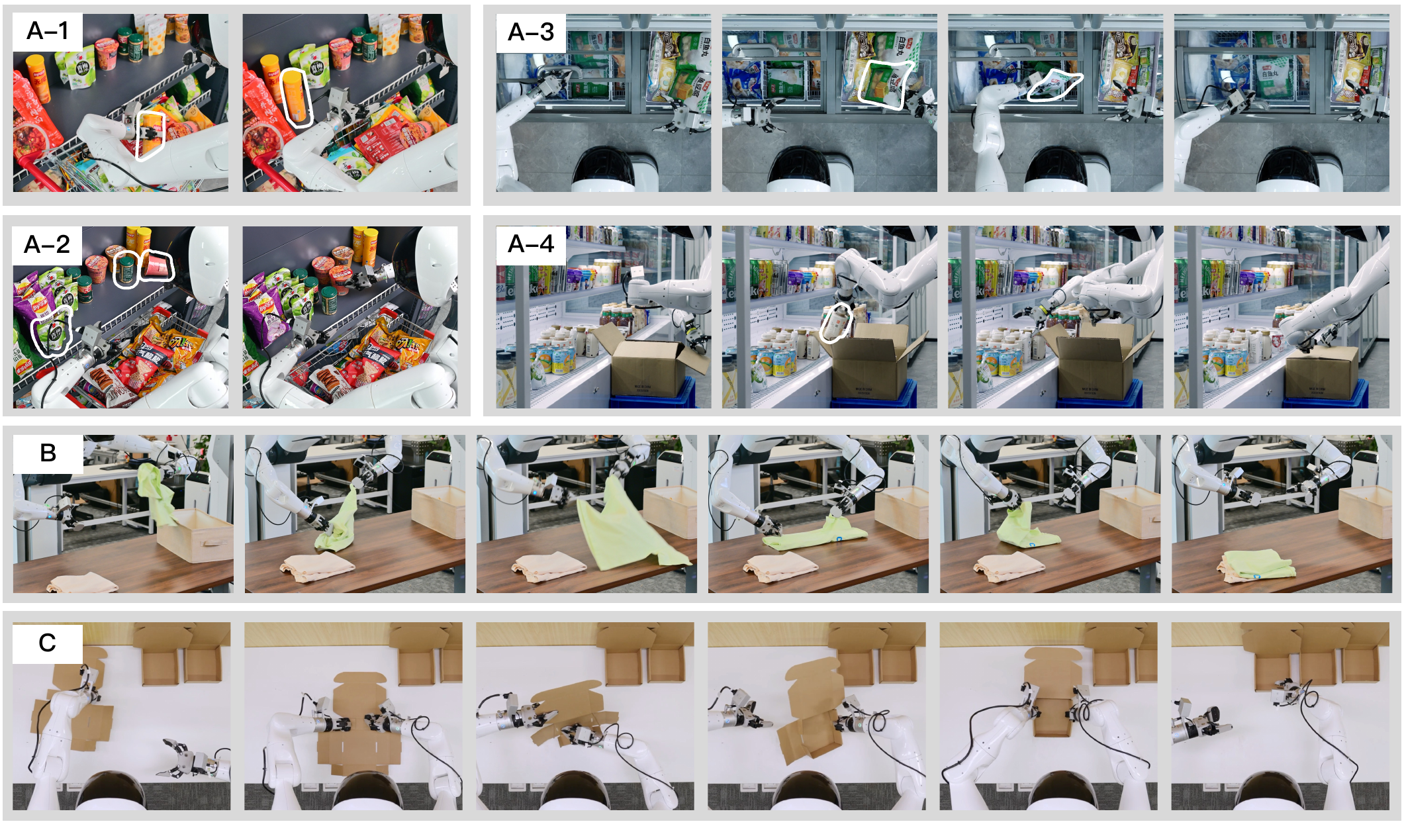}
    \caption{Illustrations of the three task categories. (A) Grocery Restocking scenarios: (A-1) flat-shelf restocking; (A-2) correcting misplaced items; (A-3) freezer restocking involving door manipulation; and (A-4) open-cooler restocking with carton handling. (B) Laundry Folding: a bimanual sequence where the robot flattens and folds a garment. (C) Box Assembly: a sequence showing two robot arms coordinating to fold a flattened cardboard sheet into a 3D box structure.}
    \label{fig:task}
\end{figure*}
\section{Experimental Evaluation}
\label{sec:experiments}
We evaluate SOP by instantiating it with two representative post-training algorithms: HG-DAgger~\cite{kelly2019hg} and RECAP~\cite{amin2025pi}. 
Experiments are conducted on a fleet of 10 dual-arm manipulators across three challenging manipulation task families as in Fig.~\ref{fig:task}, designed to stress both fine-grained dexterity and semantic generalization. Our experiments focus on three questions: 
\begin{enumerate}
    \item How effectively does SOP improve pretrained VLA performance on real-world manipulation, and how does it compare to offline alternatives?
    \item How does performance scale with fleet size?
    \item Does SOP provide consistent gains across pretrained models of varying initial quality?
\end{enumerate}
%(1) How effectively does SOP improve pretrained VLA performance on real-world manipulation, and how does it compare to offline alternatives? (2) How does performance scale with fleet size? (3) Does SOP provide consistent gains across pretrained models of varying initial quality? 
We organize results accordingly into multi-task post-training, fleet scaling, and pre-training quality/data efficiency.

\subsection{Experiment Tasks}
% \textcolor{blue}{success criteria}
We evaluate on three task families that require both dexterous manipulation and semantic understanding: Grocery Restocking, Laundry Folding, and Box Assembly. The illustration of tasks can be seen in Fig.~\ref{fig:task}.
We summarize detailed task setups and evaluation protocol below.

\textbf{Grocery Restocking.} This task evaluates policy generalization in a cluttered retail environment, requiring semantic understanding across a large catalog of store items and diverse shelf configurations. During pre-training, the robot is trained to execute diverse operations including restocking, picking, hanging, and item rearrangement. The pretraining corpus covers 500+ distinct objects in this task family. For evaluation, we uniformly sample a fixed set of 40 objects from this pool and keep it unchanged across all experiments; each of the four variants: (1) flat-shelf restocking, (2) correction for misplaced items, (3) freezer restocking with door operation, and (4) open-cooler restocking with carton handling---is tested on 10 objects drawn from this evaluation set, with 5 trials per object. During post-training, objects are sampled from the full object pool; consequently, a typical post-training session overlaps with roughly two-thirds ($\sim$70\%) of the evaluation objects, while still preserving object diversity beyond the evaluation subset. Success requires both instruction compliance (selecting the correct item from a cluttered set) and task completion (placing it at the target location within the time limit).

\textbf{Laundry Folding.} This task requires dexterous bimanual manipulation of deformable objects. During training, robots learn to pick, fold, and stack garments from a basket. For evaluation, we present a single disordered T-shirt; success requires folding it correctly and placing it on a designated stack within 500~s.

\textbf{Box Assembly.} This task evaluates precise multi-step procedural execution. The robot must transform a flat cardboard sheet into a 3D box through a sequence of folds. Success requires completing the assembly within 300~s with no folding errors.

For Laundry Folding and Box Assembly, our data collection and deployment can involve \emph{full-cycle} continuous operation (e.g., fetching a garment from a basket and repositioning it before folding; or preparing the cardboard before assembly). In quantitative evaluation, we focus on the core long-horizon manipulation skill (folding/assembly) and therefore do not include upstream fetching/stacking/preparation steps in the trial definition. Each trial begins from a randomized, disordered initial state where the garment/cardboard is placed within the robot workspace (with natural variability in pose and configuration) and ends upon success, failure, or timeout.

\vspace{0.25em}
\noindent\textbf{Metrics.}
We report \emph{success rate} (fraction of successful episodes) and \emph{throughput} (completed episodes per hour). An episode is considered \emph{completed} when it terminates due to success, failure, or timeout. Throughput therefore captures both execution speed and reliability under a fixed time budget (500~s for Laundry Folding and 300~s for Box Assembly; Grocery Restocking uses a task-dependent time limit). Importantly, throughput is measured as \emph{policy-side throughput} and does not include human operator time for environment reset or scene setup between trials; this exclusion is consistent across all compared methods.

\begin{figure*}[t!]
  \centering
  % \fbox{\parbox[c][3.5cm][c]{0.9\columnwidth}{\centering
  % Placeholder: Throughput and success-rate bar plots.}}
  \includegraphics[width=\linewidth]{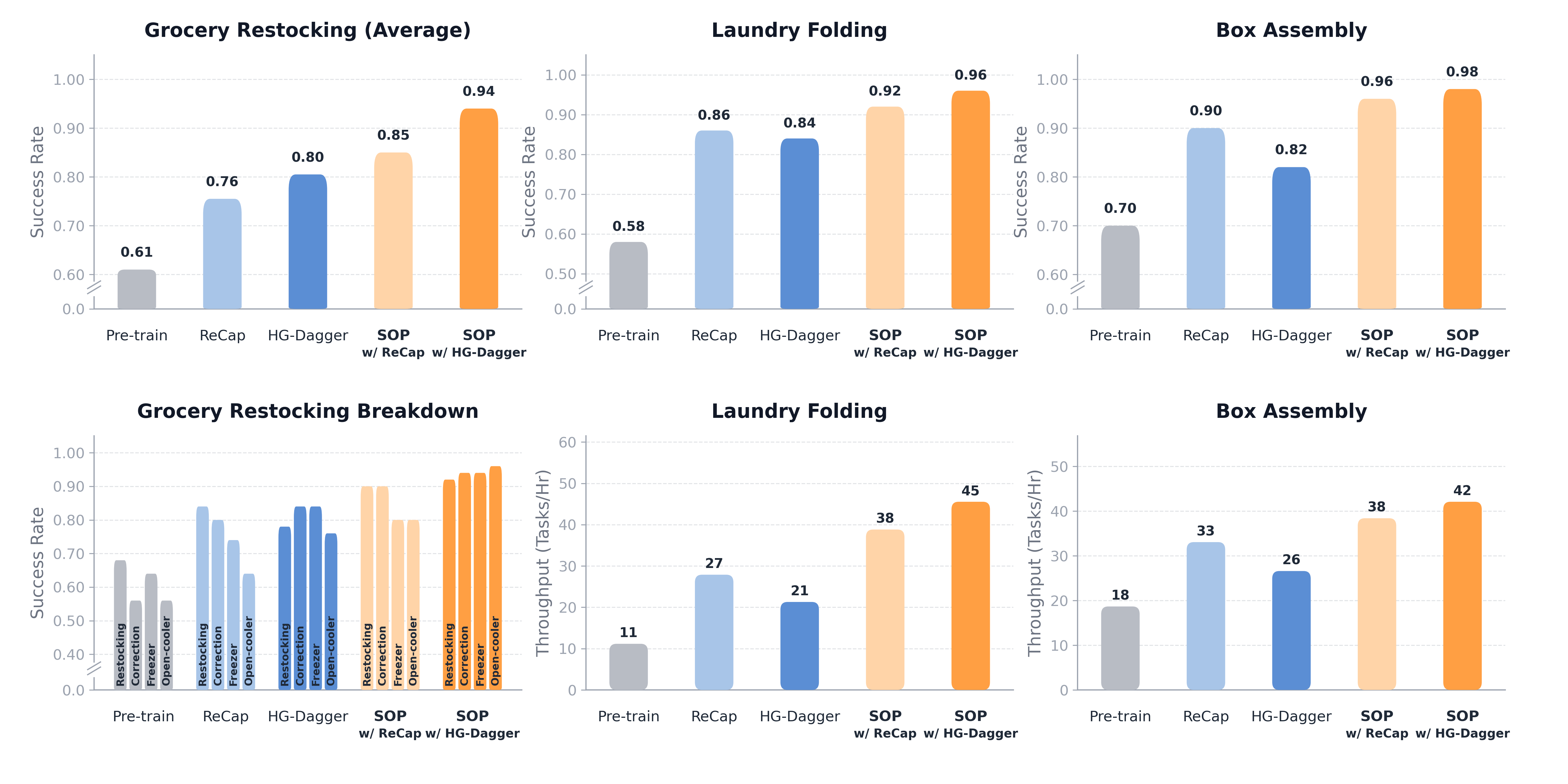}
  \caption{Comparison of Success Rate and Throughput across three manipulation domains. Across all domains, our approach demonstrates superior efficiency and reliability. SOP w/ HG-DAgger consistently achieves 2-4x higher throughput and significantly reduces failure rates compared to offline methods under our evaluation protocol (policy-side throughput excluding human reset/setup time).}
  \label{fig:exp_main}
\end{figure*}

\subsection{Experiment Setup}
We evaluate SOP with a fleet of Agibot G1 dual-arm manipulators across three task families (Grocery Restocking, Laundry Folding, and Box Assembly). In the main multi-task setting, we train a single shared learner while partitioning the 10-robot actor fleet across tasks: 4 robots collect on-policy experience for Grocery Restocking, 3 for Laundry Folding, and 3 for Box Assembly; experience from all actors is aggregated for joint SOP training.
All post-training experiments start from a pretrained base policy $\pi_{\theta_0}$, which is initialized by $\pi_{0.5}$~\cite{intelligence2025pi_}. Details are provided in the Appendix \ref{app:impl}. SOP improves it through continual on-policy interaction. Unless otherwise specified, we allocate a wall-clock budget of 3 hours (180 minutes) per experiment, and choose SOP+HG-DAgger as the post-training algorithm.
In our experiments, we train the learner on NVIDIA H100 GPUs. In 10-actor experiment setup, we provide 8 GPUs to accommodate the higher data throughput, while in other experiments we use 4 GPUs. This reflects our experimental allocation rather than a fixed requirement of SOP. In additional ablation experiments, we focus on Grocery Restocking only and use a smaller actor fleet of 4 robots, varying the number of active actors ($N \in \{1,2,4\}$). Implementation details of the pretrained base policy and SOP post-training are provided in Appendix~\ref{app:impl}.

\subsection{Multi-task Post-training}
As shown in Fig.~\ref{fig:exp_main}, we report success rate and throughput for post-trained models with and without SOP.
For Laundry Folding and Box Assembly, each reported number is evaluated over 50 trials.
For Grocery Restocking, each of the four variants is evaluated over 50 trials (10 objects $\times$ 5 trials), for a total of 200 trials across the task family.

For RECAP, the original implementation is not designed for multi-task post-training. We therefore condition both the policy and value function on the task language prompt, resulting in a multi-task variant of RECAP. To verify that this modification does not degrade performance, we additionally train single-task RECAP models on two representative grocery restocking variants: open-front cooler replenishment and freezer replenishment. Single-task RECAP achieves success rates of 0.86 and 0.75, respectively, while multi-task RECAP achieves 0.80 and 0.75 (50 trials each). These results indicate that multi-task conditioning does not significantly degrade RECAP performance, ensuring a fair comparison. Implementation details of SOP + RECAP are provided in the Appendix.

Across all tasks, post-trained models consistently outperform the pretrained baseline. Moreover, combining either HG-DAgger (or RECAP) with SOP yields substantially higher performance than their non-SOP counterparts. SOP + HG-DAgger achieves the strongest result—success rates of 0.94, 0.96, and 0.98 across the three task families. In grocery restocking, the performance gap between SOP + HG-DAgger and SOP + RECAP is the largest, which is expected given the strong semantic generalization required by this task. In such settings, learning an accurate value function with sufficiently broad coverage remains challenging, whereas interactive imitation benefits more directly from corrective supervision.
In terms of throughput, SOP substantially improves performance across all pretrained models, typically by approximately 2×. This gain arises from prompt on-policy correction that directly targets the failure modes of the deployed policy. For example, in laundry folding, a common failure mode is repeated missed grasps; SOP rapidly corrects this behavior through on-policy feedback, leading to significantly reduced cycle time.

Overall, these results demonstrate that SOP provides a simple yet effective system-level mechanism for post-training generalist VLA models through scalable robot deployment.

\subsection{Scaling robot deployments}
A natural question is how post-training efficiency scales with fleet size. 
We study how post-training efficiency scales with the size of the robot fleet by varying the number of active robot actors ($N \in {1, 2, 4}$) and evaluating both final performance and time-to-target. All experiments are capped at 180 minutes. We report the final success rate at the end of training as well as the wall-clock time required to first reach a target success level (set to 0.8 in our experiments), which we refer to as time-to-target and use as a measure of training efficiency.
As summarized in Table~\ref{tab:combined_results}, increasing the number of robot actors consistently improves both the achievable performance and the rate of learning. Expanding the fleet from one to four actors raises the final success rate at 180 minutes from 0.805 to 0.925. This improvement suggests that parallel data collection across multiple robots provides more diverse on-policy experience, reducing overfitting to station-specific noise and idiosyncrasies that arise in single-robot settings.
In addition to improving the performance ceiling, fleet scaling substantially accelerates learning. As summarized in Table~\ref{tab:combined_results}, time-to-target decreases from 173.6 minutes with a single actor to 126.5 minutes with two actors (1.4$\times$ faster) and to 71.7 minutes with four actors (2.4$\times$ faster). Within the tested regime ($N\in\{1,2,4\}$), this suggests approximately linear wall-clock speedups, indicating that SOP largely translates robot parallelism into faster on-policy post-training rather than being bottlenecked by centralized learning or communication overhead.

Overall, these results demonstrate that SOP enables post-training efficiency to scale favorably with fleet size, improving both data efficiency and wall-clock training time. They further suggest that, for real-world VLA post-training, scaling robot deployments can be as impactful as algorithmic refinements in accelerating learning.

% We vary the number of active robot actors ($N \in \{1, 2, 4\}$) and evaluate both final performance and time-to-target. As summarized in Table~\ref{tab:combined_results}, scaling out improves the final success rate and reduces the wall-clock time required to reach the target success level (time-to-target).\ly{add res of one-actor with 180x4 training time, which has similar training data with 4 actors with 180 training times in the appendix}

% All experiments are capped at 180 minutes (3 hours). We report (i) the final success rate at 180 minutes and (ii) the wall-clock time to first reach the target success level (time-to-target), which we use to quantify training efficiency.

% % --- 精修全线表 ---
% \begin{table}[htbp]
% \centering
% \renewcommand{\arraystretch}{1.3}
% \caption{Scalability with actor count ($N$). Reported success rate gains and training time reductions are computed relative to the $N{=}1$ setting.}
% \label{tab:combined_results}
% \begin{tabular}{l|c|c|c}
% \hline
% \textbf{Actor Number ($N$)} & \textbf{1} & \textbf{2} & \textbf{4} \\ \hline
% Final Success Rate @180min (\%)  & 80.5      & 88.7      & \textbf{92.5} \\ \hline
% $\Delta$Success vs. $N{=}1$ (pp) & ---       & +8.2      & \textbf{+12.0} \\ \hline \hline
% Time-to-target (mins)            & 173.56    & 126.49    & \textbf{71.66} \\ \hline
% Time Reduction vs. $N{=}1$ (\%)  & ---       & 1.4x faster    & \textbf{2.4x faster} \\ \hline
% \end{tabular}
% \end{table}

\begin{table}[t]
\centering    
\caption{Scalability Analysis with actor count. Comparison of final success rates and training efficiency scaling with different actor counts ($N$). Performance gains and speedups are reported relative to the single-actor baseline. Time-to-target denotes the wall-clock time to first reach the target success level (0.8).}
\label{tab:combined_results}

\begin{tabular}{c| l l}
\toprule
\textbf{Actor Number} & \textbf{Success Rate @180min} & \textbf{Time-to-target (min)}  \\
\midrule

$1$ & 0.805 & 173.6 \\

$2$  & 0.887 (+0.082) & 126.5 ($1.4\times$ faster)\\

$4$  & \textbf{0.925} (\textbf{+0.12}) & \textbf{71.7} (\textbf{2.4$\times$ faster}) \\

\bottomrule
\end{tabular}
\end{table}

\subsection{Analysis of Pre-training Quality and Data Efficiency}
\begin{figure}[t]
  \centering
  % \fbox{\parbox[c][3.3cm][c]{0.9\columnwidth}{\centering
  % Placeholder: Throughput vs.\ number of robots.}}
  \includegraphics[width=\linewidth]{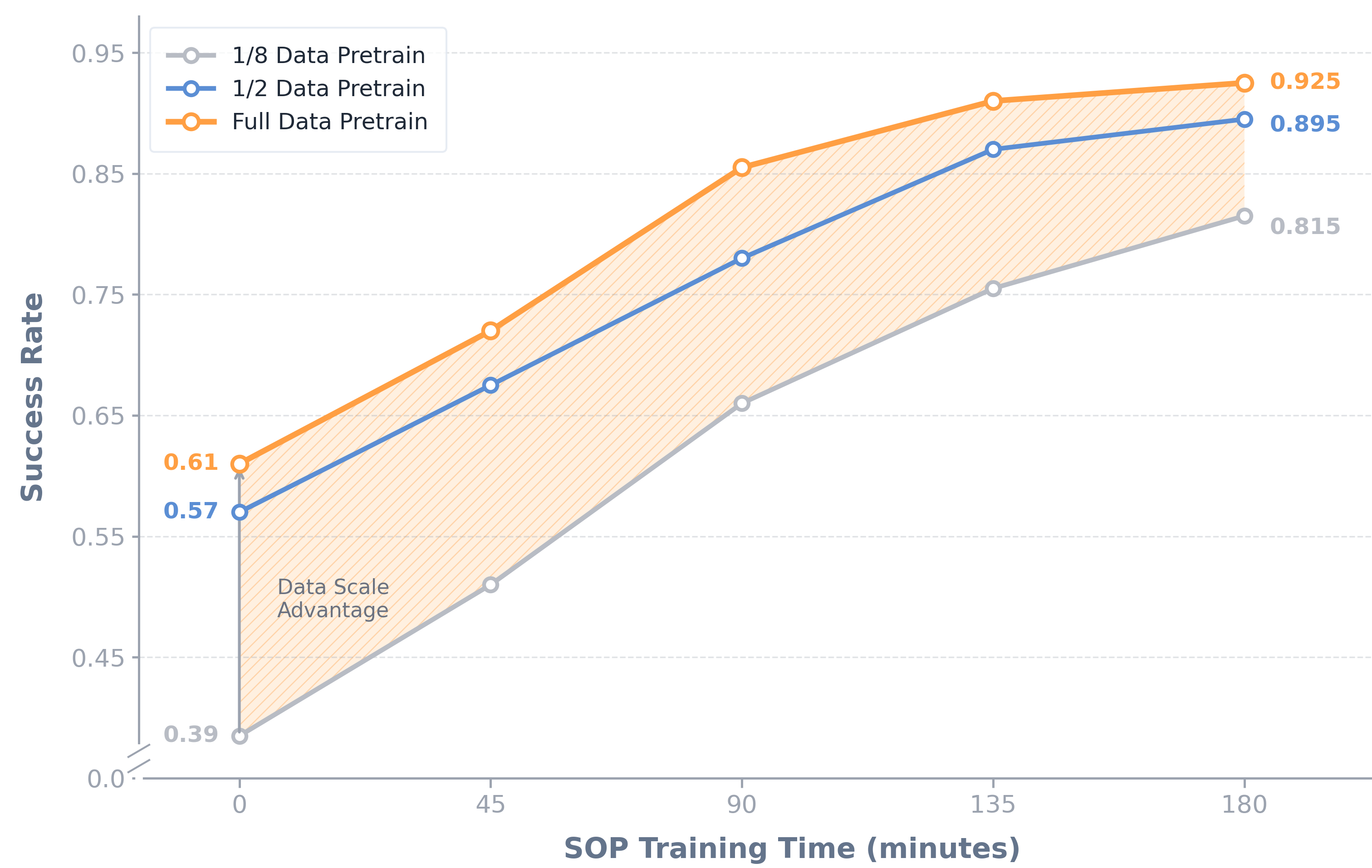}
  \caption{Effect of pretraining data scale on SOP. Larger pretraining datasets yield higher initial success and higher final performance after online post-training.}
  \label{fig:pretrain_data}
\end{figure}

To examine how the initial capacity of the base policy affects online adaptation, we evaluate SOP starting from three pretrained variants: Base-1/8, Base-1/2, and Base-Full. These models share the same architecture but are pretrained on 1/8, 1/2, and the full set of diverse multi-task pretraining data, respectively, where the full dataset comprises approximately 160 hours of data across all tasks. Results are summarized in Fig.~\ref{fig:pretrain_data}.

Across all settings, SOP consistently improves performance relative to the pretrained baseline. However, the final performance achieved after post-training remains strongly coupled to the scale of pretraining. Models initialized from larger pretraining datasets not only start from higher baselines but also converge to higher asymptotic performance. This suggests that large-scale pretraining provides essential representational priors that SOP builds upon, and that online post-training primarily refines and specializes existing knowledge rather than replacing the need for broad pretraining.

The contrast between offline data scaling and on-policy experience further highlights this effect. For the Base-1/2 model, augmenting the offline demonstration dataset with an additional 80 hours of human-collected data results in only a modest improvement in success rate, from 0.576 to 0.612. In comparison, applying SOP yields a substantially larger improvement—from 0.571 to 0.800—using only 3 hours of on-policy interaction. This gap reflects the diminishing returns of static expert demonstrations, which cannot anticipate the specific error distributions induced by a deployed policy. By directly collecting and learning from policy-generated failures, SOP allocates learning capacity to the most relevant regions of the state–action space, exhibiting a markedly more favorable scaling behavior for closing the final performance gap.

% \subsection{Analysis}
% %%J.L.1.4 this is where you put the state visitation thing if we have it

\section{Discussion and Future Work}

Our results suggest that the system-level coupling between execution and learning is as critical to post-training success as the underlying algorithm. By enabling robot fleets to continuously stream on-policy experience and receive updated policies in return, SOP transforms episodic fine-tuning into scalable, closed-loop learning. The observation that on-policy correction yields substantially greater marginal utility than additional offline data echoes a recurrent theme: static datasets cannot fully anticipate the state distribution induced by a deployed policy~\cite{ross2011reduction}. SOP operationalizes this insight at system scale.

Despite its effectiveness, SOP currently relies on human interventions or task-specific rewards; reducing this supervisory burden through learned reward models or foundation-model-based success detection remains an important direction. Whether near-linear scaling extends to significantly larger fleets, and how to support continual acquisition of new skills without catastrophic forgetting, are open questions.

Looking forward, we envision fleets of robots jointly maintaining a shared, ever-improving policy through deployment experience. In this view, scaling robot deployments becomes a form of scaling compute for learning---each additional robot accelerates policy improvement. Realizing this vision will require advances across systems, algorithms, and human-robot interaction, but the results here suggest that the foundations are within reach.

% \section{Conclusion}
% \mj{TODO}
% We presented a Scalable Online Post-training (SOP) system for large VLA models deployed in physical world. By continuously streaming on-policy experience and human interventions from distributed robots to the learner, and asynchronously deploying updated policies back to the fleet, SOP enables a online, distributed and multi-task learning simultaneously. Initiating SOP with HG-DAgger and RECAP, we demonstrate a remarkable performance improvement within hours of training, showing consistent gains across diverse real-world manipulation tasks while preserving generality. More broadly, we view SOP as a stronger post-training paradigm for generalist robot policies: SOP reframes post-training as continual learning through ongoing interaction with the deployment environment. As the future work, SOP lays the system foundation for large-scale real-robot RL post-training, where effective reward learning and distribution exploration can be realized as interaction scale grows.
% % We present a distributed online post-training framework for large Vision-Language-Action models operating on real robot fleets. By coupling human-corrected on-policy data with multi-task distributed execution, the system continuously refines both generalist reasoning and task-specific competence. This paradigm transforms VLA deployment from a static endpoint into a continual learning process—one where each robot’s experience contributes to collective improvement.

\bibliographystyle{plainnat}
\bibliography{references}

\appendix
\subsection{Robot Platform Setup}
\begin{figure}[h]
    \centering
    \includegraphics[width=\linewidth]{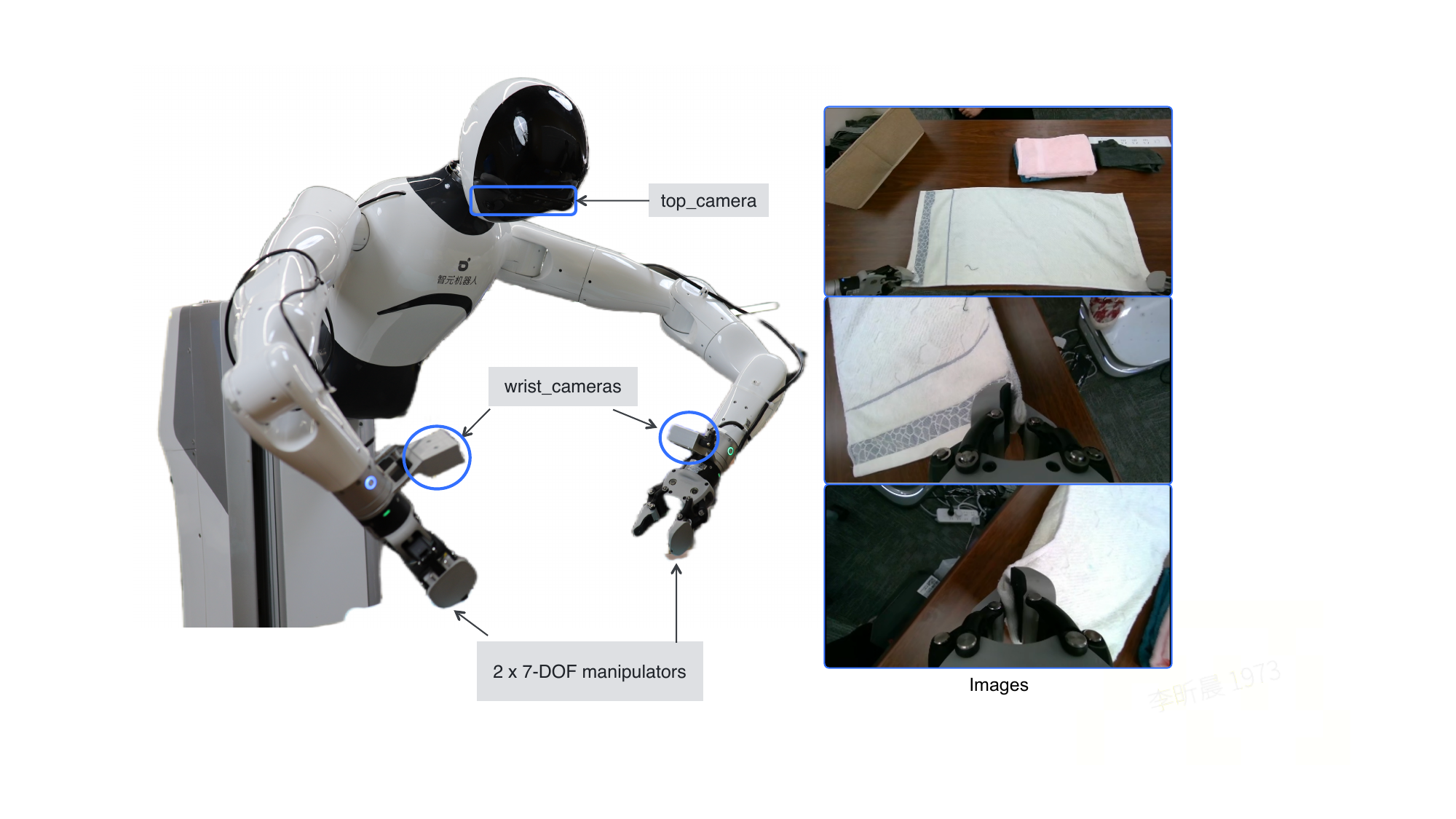}
    \caption{Robot platform setup in our experiments.}
    \label{fig:hardware}
\end{figure}

Our robot platform for experiments and evaluations is Agibot G1 dual-arm manipulator. Each G1 robot has two 7-DoF arms with parallel-jaw grippers and three RGB cameras (one base view mounted on the top head of the G1 robot and two wrist views mounted above the grippers), as illustrated in Figure \ref{fig:hardware}. Our policy executes joint position control at 30 Hz.

\subsection{Data Infrastructure Details}
\label{app-infra}

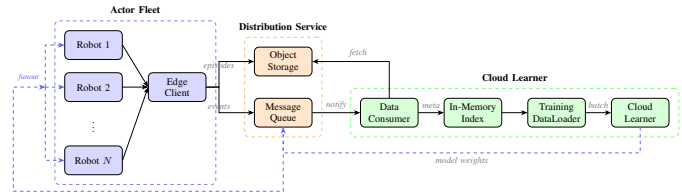
\begin{figure}[h]
    \centering
    \resizebox{\linewidth}{!}{%
    \begin{tikzpicture}[
        node distance=0.8cm and 1.2cm,
        box/.style={rectangle, draw, rounded corners, minimum height=0.9cm, minimum width=1.8cm, align=center, font=\small},
        actor/.style={box, fill=blue!15},
        distservice/.style={box, fill=orange!20},
        learner/.style={box, fill=green!15},
        arrow/.style={-{Stealth[length=2mm]}, thick},
        dashedarrow/.style={-{Stealth[length=2mm]}, thick, dashed},
        label/.style={font=\footnotesize\itshape, text=gray}
    ]

    % Actor Side
    \node[actor] (robot1) {Robot 1};
    \node[actor, below=0.4cm of robot1] (robot2) {Robot 2};
    \node[below=0.3cm of robot2, font=\small] (dots1) {$\vdots$};
    \node[actor, below=0.3cm of dots1] (robotN) {Robot $N$};

    \node[actor, right=0.8cm of robot2] (edgeclient) {Edge\\Client};

    % Distribution Service
    \node[distservice, right=1.5cm of edgeclient, yshift=0.8cm] (oss) {Object\\Storage};
    \node[distservice, right=1.5cm of edgeclient, yshift=-0.8cm] (mq) {Message\\Queue};

    % Learner Side
    \node[learner, right=1.5cm of mq] (consumer) {Data\\Consumer};
    \node[learner, right=0.8cm of consumer] (index) {In-Memory\\Index};
    \node[learner, right=0.8cm of index] (dataloader) {Training\\DataLoader};
    \node[learner, right=0.8cm of dataloader] (learner) {Cloud\\Learner};

    % Grouping boxes
    \begin{scope}[on background layer]
        \node[draw=blue!50, dashed, rounded corners, fit=(robot1)(robotN)(edgeclient), inner sep=0.3cm] (actorbox) {};
        \node[draw=orange!50, dashed, rounded corners, fit=(oss)(mq), inner sep=0.3cm] (distservicebox) {};
        \node[draw=green!50, dashed, rounded corners, fit=(consumer)(learner), inner sep=0.3cm] (learnerbox) {};
    \end{scope}

    % Group labels
    \node[font=\small\bfseries, above=0.1cm of actorbox] {Actor Fleet};
    \node[font=\small\bfseries, above=0.1cm of distservicebox] {Distribution Service};
    \node[font=\small\bfseries, above=0.1cm of learnerbox] {Cloud Learner};

    % Arrows - Actor to Distribution Service
    \draw[arrow] (robot1.east) -- (edgeclient.west);
    \draw[arrow] (robot2.east) -- (edgeclient.west);
    \draw[arrow] (robotN.east) -- (edgeclient.west);

    \draw[arrow] (edgeclient.east) -- ++(0.4,0) |- node[label, pos=0.25, above]{episodes} (oss.west);
    \draw[arrow] (edgeclient.east) -- ++(0.4,0) |- node[label, pos=0.25, below]{events} (mq.west);

    % Arrows - Distribution Service to Learner
    \draw[arrow] (mq.east) -- node[label, above]{notify} (consumer.west);
    \draw[arrow] (consumer.north) |- node[label, pos=0.7, above]{fetch} (oss.east);
    \draw[arrow] (consumer.east) -- node[label, above]{meta} (index.west);
    \draw[arrow] (index.east) -- (dataloader.west);
    \draw[arrow] (dataloader.east) -- node[label, above]{batch} (learner.west);

    % Model sync - feedback loop (fanout multicast)
    % Path: Learner -> MQ -> below actor fleet -> left side -> fanout to all robots
    \coordinate (mqdown) at ($(mq.south) + (0,-2)$);
    \coordinate (fanout) at ($(robot2.west) + (-0.6,0)$);
    \coordinate (fanout_left) at ($(fanout) + (-1,0)$);
    \coordinate (leftbottom) at (fanout_left |- mqdown);

    % Learner to MQ (aligned entry point)
    \draw[dashedarrow, blue!60] (learner.south) -- ++(0,-0.8) -| node[label, pos=0.25, below]{model weights} (mq.south);

    % MQ down -> left below actor fleet -> up to fanout on left side
    \draw[dashedarrow, blue!60] (mq.south) -- (mqdown) -- (leftbottom) -- (fanout_left) -- (fanout);

    % Fanout to each robot (from left side)
    \draw[dashedarrow, blue!60] (fanout) -- (robot2.west);
    \draw[dashedarrow, blue!60] (fanout) |- (robot1.west);
    \draw[dashedarrow, blue!60] (fanout) |- (robotN.west);

    % Fanout label
    \node[font=\scriptsize\itshape, text=blue!60, above left=0.1cm of fanout] {fanout};
    \end{tikzpicture}%
    }
    \caption{Distributed data infrastructure architecture. Robot actors upload episodes to object storage and publish event notifications to a message queue. The cloud learner consumes notifications, retrieves episode data, and streams updated model parameters back to all actors via the publish--subscribe channel.}
    \label{fig:infra-arch}
\end{figure}

\subsubsection{System Architecture}
Our distributed data infrastructure implements a closed-loop actor--learner architecture comprising five core components, as illustrated in Figure~\ref{fig:infra-arch}.

On the actor side, each robot runs an \emph{edge client} responsible for buffering frame-level observations locally and assembling them into complete episodes. Upon episode termination, the client serializes the episode data and uploads it to a \emph{distributed object storage} layer (compatible with S3-like interfaces), while simultaneously publishing an event notification to a \emph{message queue}.

On the learner side, a \emph{data consumption service} subscribes to the message queue, retrieves newly uploaded episodes from object storage, and expands them into frame-level metadata entries in an in-memory index. A \emph{training dataloader} then samples from this index according to configurable strategies and fetches the corresponding payload data on demand.

To complete the feedback loop, the learner broadcasts updated model parameters back to all actors through a publish--subscribe channel, enabling actors to refresh their local policies at episode boundaries without interrupting ongoing data collection.

This architecture cleanly separates the data production pipeline (actors) from the data consumption pipeline (learner), allowing each to scale independently. The message queue serves as the decoupling layer that absorbs transient load imbalances and network disruptions.

\subsubsection{Design Advantages}
The above architecture provides three key advantages for large-scale online post-training.

\textbf{Elastic Horizontal Scaling.}
A key design principle is zero-configuration scalability. New robot actors can join the data collection fleet by simply connecting to the message queue—no code modifications or system reconfiguration required. The cloud learner automatically discovers and consumes data from all active actors through consumer groups that provide native load balancing. This enables seamless scaling from single-robot experiments to fleets of hundreds of robots without architectural changes.

%\noindent \textbf{Edge-Cloud Decoupled Collaboration.}

\textbf{Persistent and Fault-Tolerant Data Management.}
All episode data are durably persisted to distributed object storage with atomic write semantics—either an entire episode is successfully stored, or the operation is rolled back entirely. The message queue provides guaranteed delivery with automatic retry mechanisms, ensuring no data loss even under network partitions or node failures. This reliability is critical for long-running training campaigns where data integrity directly impacts model quality.

\textbf{Separation of Metadata and Payload.}
To support million-scale episode sampling efficiently, we decouple lightweight metadata (episode identifiers, frame counts, sampling weights) from heavy payload data (images, sensor readings). Metadata resides in memory to enable high-frequency sampling decisions, while actual frame data are lazily loaded from object storage only when selected for training. This design reduces memory footprint by over two orders of magnitude compared to full data loading, making it feasible to maintain replay buffers spanning millions of frames on commodity hardware.

\subsection{Implementation Details}
\label{app:impl}

\subsubsection{Pre-trained Base Policy}
All SOP post-training experiments initialize from a pretrained vision-language-action (VLA) base policy, denoted as $\pi_{\theta_0}$. Here, $\pi_{\theta_0}$ is obtained by tuning $\pi_{0.5}$ model on our multi-task robot dataset (about 160 hours in total: 100 hours for Grocery Restocking, 30 hours for Laundry Folding, and 30 hours for Box Assembly). Unless otherwise specified, we use \textbf{Base-Full} ($\pi_{\theta_0}$ pretrained on our full dataset) as the initialization for the main experiments.

\subsubsection{SOP Training Details}
\label{subsec:sop detail}
Post-training is run with a centralized cloud learner using 8 NVIDIA H100 GPUs. The learner publishes updated model parameters every 25 training steps, and robot actors refresh their local policies using the latest published checkpoint. During post-training, we freeze the LLM backbone and train the vision components and action experts. To keep deployment practical, we only distribute the necessary updated weights; the transmitted checkpoint artifact is about 780~MB.

\subsubsection{RECAP Implementation Details}
To achieve a generalist policy, we implement RECAP in a multi-task post-training recipe rather than as a task-specific fine-tuning method.
The policy improvements of RECAP follows
\begin{equation} \label{eq:recap_sample}
\hat{\pi}(a \mid s) \propto \pi_{\text{ref}}(a \mid s) \left( \frac{\pi_{\text{ref}}(a \mid I, s)}{\pi_{\text{ref}}(a \mid s )} \right)^\beta
\end{equation}
where $\pi_{\text{ref}}$ is the behavior policy from the collected dataset and $I = \mathbf{1}\left(A^{\pi_{\text{ref}}}(s_t, a_t) > \epsilon\right)$ is the advantage condition approximated by value function. In our multi-task setting, different advantage thresholds $\epsilon$ are imposed since the episode lengths varies across different tasks.

For the SOP-RECAP experiment, online training is following the training setup as mentioned in ~\ref{subsec:sop detail} where only the vision components and action expert are updated. Value function is pretrained offline and is not being updated along policy training.
Applying SOP with RECAP, the policy executed by actor ends in Algorithm \ref{alg:distributed_vla} accords to \eqref{eq:recap_sample}, and the policy parameters is updated according to the training objective in RECAP.
For the RECAP-alone experiment, we execute 2 iterations, achieve a performance improvement on the multi-task setting, as presented in Fig.~\ref{fig:exp_main}. 
The parameter $\beta$ during rollout data collection phase and online inference phase is set to 1.0; in the evaluation phase, moderate settings $\beta \in [1.5, 2.5]$ are adopted.

To validate the effectiveness of our RECAP implementation across multi-tasks, ablation studies are made for 2 single tasks (freezer restocking and open-cooler restocking) in the Grocery Restocking category. On the freezer restocking task, the success rate of single-task RECAP vs multi-task RECAP is 0.75 vs 0.75; on the open-cooler restocking task, the success rate is 0.86 vs 0.8.

\end{document}